\documentclass[final,times,sort&compress]{elsarticle}

\usepackage{lineno,hyperref}
\modulolinenumbers[5]

\usepackage{longtable}
\usepackage{tabulary}
\usepackage{graphicx}

\usepackage{subfig}
\usepackage [english]{babel}
\usepackage [autostyle, english = american]{csquotes}
\MakeOuterQuote{"}

\usepackage{listings}
\usepackage{enumerate} 
\usepackage[section]{placeins}
\usepackage{amsmath}

\usepackage{booktabs}
\usepackage{multirow}
\usepackage{float}

\usepackage{fancyhdr}
\usepackage{textcomp}

\usepackage{threeparttable}

\fancypagestyle{plain}{%
  \fancyhf{}
  \fancyfoot[C]{\iffloatpage{}{\thepage}}
  }
\journal{a Journal}
 
\begin{document}

\begin{frontmatter}	

\title{Care2Vec: A Deep learning approach for the classification of self-care problems in physically disabled children}


%

\author[myu]{Sayan Putatunda \corref{cor1}}
\ead{sayanp@iima.ac.in}

\address[myu]{Indian Institute of Management Ahmedabad, India}

\cortext[cor1]{Tel.: +91-079-66322920}

\begin{abstract}
Accurate classification of self-care problems in children who suffer from physical and motor affliction is an important problem in the healthcare industry. This is a difficult and a time consumming process and it needs the expertise of occupational therapists. In recent years, healthcare professionals have opened up to the idea of using expert systems and artificial intelligence in the diagnosis and classification of self care problems. In this study, we propose a new deep learning based approach named Care2Vec for solving these kind of problems and use a real world self care activities dataset that is based on a conceptual framework designed by the World Health Organization (WHO). Care2Vec is a mix of unsupervised and supervised learning where we use Autoencoders and Deep neural networks as a two step modeling process. We found that Care2Vec has a better prediction accuracy than some of the traditional methods reported in the literature for solving the self care classification problem viz. Decision trees and Artificial neural networks. 
\end{abstract}


\end{frontmatter}


\section{Introduction}
\label{intro}
It is difficult to get a consensus on a proper definition for disability \citep{care:5}. Halfon et al. \cite{care:5} defined diability as a limitation instead of a health or mental issue, which restricts an individual in performing and participating in various activities desired by the society. This definition buttresses the relationship between functioning, health and the environment instead of just focusing on the common notion of some particular causes of individual disabilities such as autism, cystic fibrosis, attention-deficit/hyperactivity disorder (ADHD) and others. There have been some frameworks developed for classification of disability. One such popular conceptual framework is the "International Classification of Functioning, Disability, and Health for Children and Youth (ICF-CY)" \citep{icf:1,icf:2,icf:3,icf:4}. The World Health Organization (WHO) endorsed "International Classification of Functioning, Disability and Health (ICF)" in \cite{icf:9} which laid the foundation for ICF-CY. The ICF-CY focuses more on the bodily functions, participation, activities and environment specific to infant and children i.e. for youths from birth to the age of 18. The purpose of ICF-CY is to help doctors, educators and policy makers to define and document the development, health and functioning of children \citep{icf:8}. So, the ICF-CY can help profile a child by defining his functional / health issues and finally assigning some codes to it. This conceptual framework has thus great applications in medicine, policy, surveillance and research \citep{icf:8}. The self-care activities in children come under the participation component of ICF-CY and the classification of self care problems in children using ICF-CY can help in the appropriate choice of treatment approach \citep{care:1}.

The disability diagnosis problem is considered as a difficult problem to solve in the medical literature and need the help of expert therapists without whom the treatment becomes more difficult \citep{care:2}. The literature is replete with works that have used expert systems for disability diagnosis and classification \citep{care:1,care:7,care:8}. We discuss in details on some of these seminal works in the literature in Section \ref{lit}. These expert systems help the therapists in faster diagnosis and thus results in effective treatment. Most of these expert systems are advanced machine learning algorithms viz. Artificial neural networks (ANN) and Decision trees. 

In this paper, we focus on the self care classification problem and propose a new methodology for solving it. Deep learning is a form of "representation learning" that is used to extract high level features from the given raw data input \citep{dl:1}. It has been very successful in many applications such as computer vision \citep{dl:2}, speech recognition \citep{dl:3} and natural language processing \citep{dl:4}. The main contribution of this paper is in development of a new method "Care2Vec", which is a deep learning approach for solving the classification of self-care problems in children who suffer from physical and motor affliction. We then compare the proposed method with traditional methods used in the literature viz. Artificial neural networks and Decision trees. We found that Care2Vec has a better prediction accuracy than the prevalent methods for solving self care classification problem and this can help expert therapists in making better diagnostic decisions. This will thus lead to better treatment.

The rest of this paper is organized as follows. Section \ref{lit} gives a brief review of literature which is followed by Section \ref{mat} that describes the data used for experiments in this paper along with a discussion on the relevant machine learning algorithms and the proposed methodology. Section \ref{exp} presents the results of the various experiments conducted in this paper, which is followed by a discussion in Section \ref{discuss}. Finally, Section \ref{conclusion} concludes the paper.

\subsection{Related Work} \label{lit}
There have been many works in the literature on applications of machine learning and expert systems for classification and identification of disabilities. Wu et al. \cite{care:7} worked on identification of students with learning disabilities that includes dyslexia, attention deficit disorder and more using ANN. They recommend their model as "second opinion" to experts who in charge of learning disabilities evaluation. Hofmeister and Lubke \cite{care:10} and Rasli et al. \cite{care:11} are some other works on applications of Artificial intelligence (AI) for diagnosis and classification of learning disabilities. 

Rajkumar et al. \cite{care:12} worked on an intelligent decision support system for the evaluation of the degree of hearing loss. Wu et al. \cite{care:13} worked on learning the behavior patterns of people with disabilities using machine learning methods such as Hidden markov models. Varol et al. \cite{care:14} implemented multiple machine learning algorithms viz. Decision trees, Support vector machine and k nearest neighbors for identification of children with reading disability. 

Zarchi et al. \cite{care:1} worked on the self-care identification and classification problem in children and they came up with an innovative and the first of its kind dataset for this purpose known as "SCADI (Self-Care Activities Dataset based on ICF-CY)". They used advanced machine learning algorithms viz. Artificial neural networks and Decision trees for solving the self-care problem classification. 

However, in the literature we don't find much work apart from that of Zarchi et al. \cite{care:1} that focuses on machine learning algorithms for solving the self-care problem classification. Thus, there is a great opportunity for development and application of new methodologies that outperform the traditional ones in this domain.

\section{Material and methods} \label{mat}
In this section, we discuss the various materials and methods used in this paper. Section \ref{data} describes the dataset used for our experiments. In Section \ref{tree}, \ref{ann} and \ref{dl}, we give a brief background on Decision trees, Artificial neural networks and Deep learning respectively. In Section \ref{care2vec}, we describe our proposed methodology i.e. Care2Vec.

\subsection{Data} \label{data}
We use the publicly available dataset "SCADI (Self-Care Activities Dataset based on ICF-CY)" \citep{uci:1} for our experiments in this paper. The SCADI dataset was first used in Zarchi et al. \citep{care:1}. SCADI is the only standard dataset publicly available so far that can be used for studying the self care activities in children who suffer from physical and motor affliction \citep{care:1}. 
This dataset was created in collaboration with experienced occupational therapists and comprises features such as chidren age, gender, self care activity codes and target class for $70$ children. 

The SCADI dataset considers $29$ self care activities such as looking after one's health and safety, toileting, eating, drinking, dressing, washing oneself and more. Each of these $29$ self care activities have $7$ different ICF-CY codes citing the extent of impairment \citep{icf:8}. Thus the dataset has a total of $203$ self care activity features (i.e. after converting them into binary variables) along with age and gender. 

In the dataset, the expert therapists identified and categorized the self care problems into $7$ groups in the target class viz. (a) issue with caring for body parts, (b) issue with toileting, (c) issue with dressing, (d) issues with washing oneself, caring for body parts and dressing, (e) issues with toileting, washing oneself, caring for body parts and dressing, (f) issues with eating, drinking, toileting, washing oneself, caring for body parts and dressing and (g) no issues (Please see Zarchi et al. \citep{care:1} for more details on the SCADI dataset). In the SCADI dataset, the above mentioned groups from (a) to (g) are referred as Class1, Class2, Class3, Class4, Class5, Class6 and Class7 respectively. The total number of observations in each of these classes are $2$, $7$, $1$, $12$, $3$, $29$ and $16$ respectively.

\subsection{Background and Methodology} \label{method}

\subsubsection{Decision trees} \label{tree}
Classification methods aid in automated/rule-based decision making. Decision trees are a supervised learning method and are simple and easy to interpret for any audience \citep{stat:4}. The first decision tree was developed by Sonquist and Morgan which was followed by CHAID \citep{stat:6} and CART \citep{stat:2}. Some of the contemporary methods such as ID3 and C4.5 are based on the Information theory \citep{stat:4}.

Decision trees can be applied to both regression and classification problems \citep{stat:2}. Prediction using decision trees is performed by stratification of the feature space i.e. first the feature space comprising $X_1,X_2,.....,X_n$ is divided into $P$ distinct and non-overlapping regions $D_1, D_2,...., D_P$. Then for every observation in the region $D_p$ a prediction is made based on majority class voting \citep{stat:1}. The classification tree is grown using recursive binary splitting with the classification error rate as the criterion. However, it has been found that the classification error rate is not sensitive for the growth of tree and so there are other measures such as the Gini index and the cross entropy are generally preferred \citep{stat:1}. See Loh \citep{stat:5} for a detailed review of decision trees.

\subsubsection{Artificial Neural Networks} \label{ann}
Neural networks are used in various machine learning and statistical modeling problems such as classification, regression and forecasting. It has great applications in various fields such as such as  Forex prediction \citep{ann:12}, weather forecasting \citep{ann:10}, Location/Travel time prediction for GPS Taxis \citep{sayan:2, thesis:sayan} and more.

Some studies such as Cheng and Titterinton \citep{ann:7} perceive neural networks as an alternative to non-linear regression. Some of the earliest works in neural networks was done by McCulloch and Pitts \citep{ann:6}, Rosenblatt \citep{ann:3}, Rumelhart et al. \citep{ann:4} and  Widrow and Hoff \citep{ann:5}. In the context of this paper and the various studies reported in the self-care classification problem literature, we discuss the simple "feed-forward neural networks" that consists of a input layer, one or more hidden layers with hidden nodes and an output layer \citep{ann:1}. 

The hidden node is like a processing element that transforms and maps the input variables to some outputs using special function known as the activation function. Each hidden node has a weight and a bias term added to it. Let us suppose that the input layer consists of $K$ nodes for the input variables $p_1,\ldots, p_K$.
\begin{equation}
q_j^{(a)} = \sum_{i=1}^{K} \; w_{ji}^{(a)} p_i \; + \; w_{j0}^{(a)}
\label{nn1}
\end{equation}
where $ j= 1, \ldots, T$ and $T$ is the number of hidden nodes. As we can see in Equation \ref{nn1} that $T$ linear combinations of the input variables are formed. In Equation \ref{nn1} the parameters belong to the first layer of the neural network, which is denoted by the superscript (a). The weights and the bias terms are represented by $w_{ji}^{(a)}$ and $w_{j0}^{(a)}$ respectively. $q_j^{a}$ are known as the activations which are transformed by a non-linear differentiable activation function $\sigma(\cdot)$ \citep{ann:1,sayan:1}. One of the most common activation function is the Sigmoid function \citep{ann:2}, but in this paper we use the ReLu activation function in the hidden layers and the Softmax function (since we are dealing with a multi-class classification problem) in the output layer. Please see Bishop \citep{ann:1} for a detailed account on Artificial neural networks.

\subsubsection{Deep Learning} \label{dl}
Deep learning can be seen as a subset of machine learning where the focus is on learning from successive layers where each layer represents some meaningful representations \citep{dl:5}. In other words, Deep learning is a form of "representation learning" that is used to extract high level features from the given raw data input \citep{dl:1}. Deep learning has been very successful in many applications such as computer vision \citep{dl:2}, speech recognition \citep{dl:3} and natural language processing \citep{dl:4}. It has also improved other applications such as search results on web, advertisement targeting, machine translation and autonomous driving \citep{dl:5}.

Artificial neural networks (see Section \ref{ann}) are used to learn the layered representations in a deep learning model. In case of a Deep neural networks (DNN), one increases the number of hidden layers, which determined the "depth" of the deep learning model \citep{dl:5}. Apart from Deep neural networks, there are other types of deep learning models such as "Convolution neural networks (CNN)" and "Recurrent neural networks (RNN)" which we won't discuss in detail in this paper. Please see Goodfellow et al. \citep{dl:1} for a detailed account of Deep learning methods. However, we will discuss Autoencoders i.e. another form of deep learning method below.

\paragraph{Autoencoders}Chollet \citep{dl:5} categorized Autoencoders into the "self-supervised learning" category since it is a supervised learning method without the label annotations. An Autoencoder comprises of an "encoder" function and a "decoder" function. The encoder function is used to convert the input raw data into various representations and the decoder function converts the representations back to the original input format. The Autoencoder preserves as much as information is possible and also add new representations with different properties when the input raw data is passed through it \citep{dl:1}. Autoencoders have great applications in various fields such as data dimensionality reduction \citep{auto:1}, cyber security \citep{auto:2}, Smartphone-based Human activity recognition \citep{auto:3} and many more.

\subsection{Care2Vec method} \label{care2vec}
\begin{figure}[!htp]
 \centering
  \includegraphics[width=0.75\textwidth]{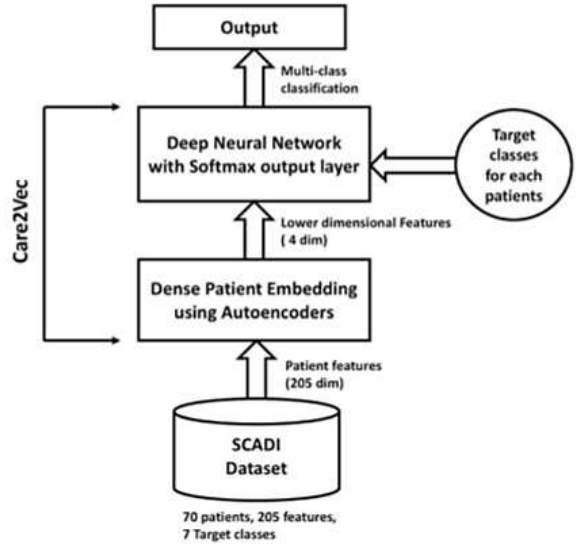}
  \caption{The Care2Vec method process flow}
  \label{fig:boat1}
\end{figure}

Figure \ref{fig:boat1} shows the modus-operandi of the Care2Vec method, which is a two-staged deep learning approach. The first step is to create a dense embedding of the patient features (in this case a child's information from the dataset mentioned in Section \ref{data}) from a high dimensional space to a low dimensional space i.e. we convert the 205 features into a low (viz. $4$, $8$, $16$ or $32$) dimensional feature set using Autoencoders (see Section \ref{dl}). In this paper, we use different encoding dimension of the autoencoder viz. $4$, $8$, $16$ and $32$. The autoencoder contains three encoder and decoder layers comprising $200$, $100$ and $50$ hidden nodes with ReLu activation function.

Figure \ref{fig:boat1} describes the care2vec algorithm for solving the self-care classification problem in a multi-class classification setting. We will also solve this problem in a binary classification setting (see Section \ref{binary}).

\section{Results} \label{exp}
In this paper we are dealing with the self-care classification problem in two different settings. In first case, we are looking into this as a multi-class classification problem since there are multiple target classes that represents the self-care problems (i.e. $7$ as described in Section \ref{data}) and have $205$ features/predictors. In the second case, we are looking at the self-care classification problem in a binary classification setting where the target class represents whether the person is suffering from any self-care problem or not. The entire model development and data analysis is implemented using Python \citep{py:1}. The libraries used for Deep learning were Keras \citep{py:2} and Tensorflow \citep{py:3}. The decision tree model was implemented using the sci-kit learn library \citep{py:4}. A system with a configuration of 16 GB RAM, 2.9 GHz Intel Core i7 processor and a 64 bit Mac OSX was used to carry out the experiments.

In this section we will first discuss the evaluation metrics (see Section \ref{eval}) and then the experimental results for both the multi-class classification and the binary classification problems in Sections \ref{multi} and \ref{binary} respectively.

\subsection{Evaluation Metrics} \label{eval}
In this paper, the primary evaluation criteria that we have used for evaluating the different models is the mean cross-validation (CV) score. We will perform the k-fold cross validation technique where the dataset is partitioned into k equal segments or folds. This is followed by $k$ subsequent iterations 	with one fold held out for validation and $k$ folds are used for training the algorithm \citep{eval:1}. We compute the prediction accuracy score for each iterations and take the average of these scores i.e. the "Mean CV score" as the evaluation metric for the performance of the method. The higher the mean CV score, the better the performance of the method.

We use the ROC area under the curve (AUC) as an evaluation metric in the binary classification setting (see Section \ref{binary}). Fawcett \cite{eval:2} defines ROC curve as a graphical depiction of a classifier's performance. The AUC of the ROC curves are computed, which represents the predictive performance of the method. In the context of binary classification, the AUC of a classifier/method represents the probability that the classifier which rank a randomly chosen instance of "class 1" higher than that of randomly chosen instance of "class 0" \citep{eval:2}. The value of AUC lies in the range of 0 to 1 and an AUC greater than 0.5 means that the classifier/method is performing better than random guessing. Thus, a higher AUC value represents a better performance of the method.

\subsection{Self-Care Classification Problem: Multi-class Classification} \label{multi}
In Section \ref{data}, we describe the SCADI dataset where the target variable has multiple classes i.e. around $7$ classes. The goal is to predict the target classes for each children given the independent variables such as gender, age and self-care activity features. We apply our proposed method Care2Vec (see Section \ref{care2vec}) with k-fold cross validation and compare its performance using the evaluation metric we have discussed earlier i.e. the Mean CV score (see Section \ref{eval}) with traditional methods such as  Decision trees and Artificial Neural Networks (ANN) that have been reported in the literature \citep{care:1}.

\begin{table}[!htp]
\centering
\caption{Mean CV score for ANN with different hidden nodes and 1 hidden layer}
\label{tab1}
 \scalebox{0.9}{
\begin{tabular}{cc} 
\hline
\begin{tabular}[c]{@{}l@{}}No. of hidden \\ nodes\end{tabular} & \begin{tabular}[c]{@{}l@{}}Mean CV \\ score (\%)\end{tabular} \\ \hline
30                                                             & 78.57                                                    \\  \hline 
40                                                             & 81.43                                                    \\  \hline 
50                                                             & 80.00                                                    \\  \hline 
100                                                            & 80.00                                                    \\  \hline 
300                                                            & 80.00        \\     \hline                                      
\end{tabular}}
\end{table}

Table \ref{tab1} describes the mean CV score for ANN for different hidden nodes. Here we can see that the ANN with $40$ hidden nodes is the best performer which is similar to what is reported in Zarchi et al. \cite{care:1}. The mean CV score for the ANN with $40$ hidden nodes is $81.43\%$.

\begin{table}[!htp]
\centering
\caption{Mean CV score for Care2Vec with different encoding dimensions in the Autoencoder, hidden nodes and hidden layers in the DNN}
\label{tab2}
 \scalebox{0.7}{
\begin{tabular}{cccc}
\hline
\multicolumn{1}{l}{\begin{tabular}[c]{@{}l@{}}Encoding \\ Dimensions\end{tabular}} & \multicolumn{1}{l}{\begin{tabular}[c]{@{}l@{}}No. of hidden \\ nodes in DNN\end{tabular}} & \multicolumn{1}{l}{\begin{tabular}[c]{@{}l@{}}No. of hidden \\ layers in DNN\end{tabular}} & \multicolumn{1}{l}{\begin{tabular}[c]{@{}l@{}}Mean CV \\ score (\%)\end{tabular}} \\ \hline
\multirow{4}{*}{4}                                                                 & 40                                                                                        & 1                                                                                          & 72.86                                                                        \\ \cmidrule(l){2-4}
                                                                                   & 100                                                                                       & 1                                                                                          & 72.86                                                                        \\ \cmidrule(l){2-4}
                                                                                   & 300                                                                                       & 1                                                                                          & 81.43                                                                        \\ \cmidrule(l){2-4}
                                                                                   & 300                                                                                       & 2                                                                                          & 81.43                                                                        \\ \hline
\multirow{4}{*}{8}                                                                 & 40                                                                                        & 1                                                                                          & 82.86                                                                        \\ \cmidrule(l){2-4}
                                                                                   & 100                                                                                       & 1                                                                                          & 81.43                                                                        \\ \cmidrule(l){2-4}
                                                                                   & 300                                                                                       & 1                                                                                          & 80.00                                                                 \\ \cmidrule(l){2-4}
                                                                                   & 300                                                                                       & 2                                                                                          & 80.00                                                                         \\ \hline       
\multirow{4}{*}{16}                                                                 & 40                                                                                        & 1                                                                                          & 80.00                                                                        \\ \cmidrule(l){2-4}
                                                                                   & 100                                                                                       & 1                                                                                          & 81.43                                                                        \\ \cmidrule(l){2-4}
                                                                                   & 300                                                                                       & 1                                                                                          & 82.86                                                                     \\ \cmidrule(l){2-4}
                                                                                   & 300                                                                                       & 2                                                                                          & 82.86                                                                        \\ \hline
\multirow{4}{*}{32}                                                                 & 40                                                                                        & 1                                                                                          & 82.86                                                                        \\ \cmidrule(l){2-4}
                                                                                   & 100                                                                                       & 1                                                                                          & 82.86                                                                        \\ \cmidrule(l){2-4}
                                                                                   & 300                                                                                       & 1                                                                                          & 80.00                                                                        \\ \cmidrule(l){2-4}
                                                                                   & 300                                                                                       & 2                                                                                          & 84.29                                                                        \\ \hline                                                                                                                                                   
\end{tabular}}
\end{table}

Table \ref{tab2} describes the mean CV score of the Care2Vec method for different hyper-parameters. The different hyper-parameters are the encoding dimensions, number of hidden layers and number of hidden nodes. As mentioned in Section \ref{care2vec}, the Care2Vec method is a two step modeling process comprising an Autoencoder and a Deep neural network (DNN). The different encoding dimensions in the autoencoder that we use in our experiment are $4$, $8$, $16$ and $32$. The different number hidden nodes we use in the DNN are $40$, $100$ and $300$ for $1$ hidden layer and for a DNN with $2$ hidden layers, we use $300$ hidden nodes in each. We find that the Care2Vec (with $32$ encoding dimensions, $2$ hidden layers and $300$ hidden nodes) method is the best performer with a mean CV score of $84.29\%$.

\begin{table}[!htp]
\centering
\caption{Mean CV score of the different methods using the best parameters}
\label{tab3}
 \scalebox{0.7}{
\begin{tabular}{cc}
\hline
Method        & \begin{tabular}[c]{@{}c@{}}Mean CV \\ score (\%)\end{tabular} \\ \hline
Decision tree & 76.99                                                    \\  \hline 
ANN           & 81.43                                                    \\ \hline 
Care2vec      & 84.29                          \\ \hline                         
\end{tabular}}
\end{table}

In Table \ref{tab3}, we summarize the results for ANN and Care2Vec with the best hyper-parameters. We also report the mean CV score when the Decision tree (with Gini criterion) is applied to the dataset. We find that the Care2Vec method is the best performer in terms of prediction accuracy compared to that of Decision tree and ANN.

\subsection{Self-Care Classification Problem: Binary Classification} \label{binary}
In this section, we handle the self-care classification problem in the binary classification setting. The target variable in the SCADI dataset (see Section \ref{data}) is modified into a binary class variable where "class7" that represents no self-care issues is marked as $1$ and all the other classes are marked as $0$. The goal is to predict the probability for each children to have no self-care issues given the independent variables such as gender, age and self-care activity features. Here also we apply our proposed method Care2Vec and compare its performance with that of traditional methods viz. Decision tree and ANN. We implement these methods using k-fold cross validation. And as far as evaluation metrics are concerned, we use AUC and Mean CV score as discussed in Section \ref{eval}.

\begin{table}[!htp]
\centering
\caption{AUC in each 10 folds, Mean AUC and Mean CV score for the different methods with different hyper-parameters viz. number of hidden nodes and hidden layers and encoding dimensions (dim)}
\label{tab4}
 \scalebox{0.53}{
  \begin{threeparttable}
\begin{tabular}{cccccccccccccc}
\hline\\
\multirow{2}{*}{Method}   & \multirow{2}{*}{hyper-parameters}                                                  & \multicolumn{10}{c}{AUC in each folds (\%)}                                                   & \multirow{2}{*}{\begin{tabular}[c]{@{}c@{}}Mean \\ AUC (\%)\end{tabular}} & \multirow{2}{*}{\begin{tabular}[c]{@{}c@{}}Mean CV \\ score (\%)\end{tabular}} \\ \cmidrule(l){3-12}
                          &                                                                              & Fold 1 & Fold 2 & Fold 3 & Fold 4 & Fold 5 & Fold 6 & Fold 7 & Fold 8 & Fold 9 & Fold 10 &                                                                      &                                                                           \\ \hline \\
Decision tree             & Gini Criterion                                                               & 65.00  & 74.24  & 75.00  & 38.46  & 74.24  & 70.83  & 70.00  & 95.83  & 80.00  & 92.31   & 73.59                                                                & 86.79                                                                     \\ \hline \\
\multirow{3}{*}{ANN}      & \begin{tabular}[c]{@{}c@{}}40 nodes, \\ 1 hidden layer\end{tabular}          & 92.50  & 96.97  & 97.50  & 84.62  & 90.91  & 95.83  & 90.00  & 100.00 & 97.78  & 100.00  & 94.61                                                                & 91.43                                                                     \\ \cmidrule(l){2-14} 
                          & \begin{tabular}[c]{@{}c@{}}100 nodes,\\  1 hidden layer\end{tabular}         & 92.50  & 96.97  & 95.00  & 84.62  & 87.88  & 95.83  & 90.00  & 100.00 & 95.56  & 100.00  & 93.83                                                                & 91.43                                                                     \\ \cmidrule(l){2-14}
                          & \begin{tabular}[c]{@{}c@{}}300 nodes, \\ 1 hidden layer\end{tabular}         & 92.50  & 96.97  & 95.00  & 84.62  & 90.91  & 95.83  & 90.00  & 100.00 & 95.56  & 100.00  & 94.13                                                                & 91.43                                                                     \\ \hline \\
\multirow{4}{*}{Care2Vec} & \begin{tabular}[c]{@{}c@{}}4 dim, 300 nodes, \\ 1 hidden layer\end{tabular}  & 92.50  & 84.85  & 82.50  & 69.23  & 96.97  & 100.00 & 82.50  & 100.00 & 100.00 & 100.00  & 90.85                                                                & 81.43                                                                     \\ \cmidrule(l){2-14}
                          & \begin{tabular}[c]{@{}c@{}}8 dim, 300 nodes, \\ 1 hidden layer\end{tabular}  & 87.50  & 93.94  & 100.00 & 92.31  & 81.82  & 91.67  & 100.00 & 100.00 & 93.33  & 84.62   & 92.51                                                                & 92.86                                                                     \\ \cmidrule(l){2-14}
                          & \begin{tabular}[c]{@{}c@{}}16 dim, 300 nodes, \\ 1 hidden layer\end{tabular} & 92.50  & 93.94  & 100.00 & 92.31  & 87.88  & 95.83  & 100.00 & 100.00 & 100.00 & 76.92   & 93.93                                                                & 90.00                                                                     \\ \cmidrule(l){2-14}
                          & \begin{tabular}[c]{@{}c@{}}32 dim, 300 nodes, \\ 1 hidden layer\end{tabular} & 95.00  & 93.94  & 100.00 & 92.31  & 93.94  & 95.83  & 92.50  & 100.00 & 100.00 & 100.00  & 96.35                                                                & 88.57               \\ \hline                                                     
\end{tabular}
\begin{tablenotes}\footnotesize
\item[*]dim= the value of encoding dimensions, nodes= number of hidden nodes in each hidden layer, hidden layer= number of hidden layers in the neural network
\end{tablenotes}
\end{threeparttable}}
\end{table}

Table \ref{tab4} describes the AUC in each 10 folds, Mean AUC and Mean CV score for the Decision tree, ANN and Care2Vec methods with different hyper-parameters. We find that the Care2Vec performs better than Decision tree and ANN taking into account both the evaluation metrics viz. Mean AUC and Mean CV score.

\section{Discussion} \label{discuss}
The ICF-CY focuses on the bodily functions, participation, activities and environment specific to infant and children i.e. for youths from birth to the age of 18. The purpose of ICF-CY is to help doctors, educators and policy makers to define and document the development, health and functioning of children \citep{icf:8}. So, the ICF-CY can help profile a child by defining his functional / health issues and finally assigning some codes to it. This conceptual framework has thus great applications in medicine, policy, surveillance and research \citep{icf:8}. The self-care activities in children come under the participation component of ICF-CY and the classification of self care problems in children using ICF-CY can help in the appropriate choice of treatment approach \citep{care:1}.

The disability diagnosis problem is considered as a difficult problem to solve in the medical literature and need the help of expert therapists without whom the treatment becomes more difficult \citep{care:2}. The literature is replete with works that have used expert systems for disability diagnosis and classification as discussed in Section \ref{lit}. These expert systems help the therapists in faster diagnosis and thus results in effective treatment. Zarchi et al. \cite{care:1} proposed some of the effective advanced machine learning algorithms viz. Artificial neural networks (ANN) and Decision trees that can be used to solve the self-care classification problem.

In this paper, we propose a new methodology for solving the self-care classification problem that hasn't been earlier reported in the medical informatics literature. We propose a deep learning approach i.e. Care2Vec, which is a two-step modeling process comprising autoencoders and deep neural networks and performs really well when applied on high dimensional datasets. We applied Care2Vec along with Decision tree and ANN on the SCADI dataset \cite{care:1} in both the multi-class classification and binary classification settings. We found that Care2Vec is the best performer in terms of prediction accuracy when compare to that of traditional approaches such as Decision tree and ANN for solving the self-care classification problem. Care2Vec can help therapists in better diagnosis and thus results in effective treatment.

\section{Conclusion} \label{conclusion}
In this paper, we focus on the self care classification problem and propose a new methodology for solving it. We developed a deep learning approach i.e. "Care2Vec" for solving the classification of self-care problems in children who suffer from physical and motor affliction. Care2Vec is a mix of unsupervised and supervised learning where we use Autoencoders and Deep neural networks as a two step modeling process. 

We compared Care2Vec with traditional methods used in the literature viz. Artificial neural networks and Decision trees for solving the self-care classification problem using the SCADI dataset in both multi-class classification and binary classification settings. We found that Care2Vec has a better prediction accuracy than the prevalent methods for solving self care classification problem. This can help expert therapists in making better diagnostic decisions and thus lead to better treatment. 

However, we feel that Care2Vec can be used in other relevant problems and different types of data in medical informatics including unstructured data such as medical reports, MRI scan images, etc. and we would be working in this direction for our future research.

\section*{Acknowledgement}
The author thanks Zarchi et al. \cite{care:1} for making the SCADI dataset publicly available without which this research work wouldn't have been possible.

\section*{References}

\bibliography{Care2Vecbib}

\end{document}